\documentclass[graybox]{svmult}

\usepackage{amsmath}
\usepackage{amsfonts}
\usepackage{amssymb}
\usepackage{mathptmx}       
\usepackage{helvet}         
\usepackage{courier}        
\usepackage{type1cm}        
\usepackage{makeidx}         
\usepackage{graphicx}        
\graphicspath{{}}
\usepackage{multicol}        
\usepackage[bottom]{footmisc}
\usepackage{acronym}
\usepackage{multirow}
\usepackage{color,soul}

\usepackage[caption=false,font=footnotesize]{subfig}
\usepackage{setspace}

\makeindex             

\newcommand{\secPref}{Sect. }
\newcommand{\figPref}{Fig. }
\newcommand{\tabPref}{Table }

\DeclareMathAlphabet{\mathcal}{OMS}{cmsy}{m}{n}
\SetMathAlphabet{\mathcal}{bold}{OMS}{cmsy}{b}{n}

\begin{document}
\acrodef{vc}[VC]{Vapnik-Chervonenkis}
\acrodef{knn}[$ k $NN]{$ k $ nearest neighbors}
\acrodef{lbs}[LBS]{location-based service}
\acrodef{ilbs}[ILBS]{indoor location-based service}
\acrodef{fips}[FIPS]{fingerprinting-based indoor positioning system}
\acrodef{gnss}[GNSS]{global navigation satellite system}
\acrodef{ips}[IPS]{indoor positioning system}
\acrodef{rfid}[RFID]{radio frequency identification}
\acrodef{uwb}[UWB]{ultra wideband}
\acrodef{wlan}[WLAN]{wireless local area network}
\acrodef{rss}[RSS]{received sigal strength}
\acrodef{ap}[AP]{access point}
\acrodef{roi}[RoI]{region of interest}
\acrodef{lasso}[LASSO]{least absolute shrinkage and selection operator}
\acrodef{rfm}[RFM]{reference fingerprint map}
\acrodef{map}[MAP]{maximum a posteriori}
\acrodef{cpa}[CPA]{cumulative positioning accuracy}
\acrodef{mse}[MSE]{mean squared error}
\acrodef{tp}[TP]{test position}
\acrodef{wrt}[w.r.t.]{with respect to}
\acrodef{mac}[MAC]{media access control}
\acrodef{imu}[IMU]{inertial measurement unit}
	
	\title*{Jaccard analysis and LASSO-based feature selection for location fingerprinting with limited computational complexity}
	\titlerunning{Location fingerprinting with limited computational complexity}
	\author{Caifa Zhou, Andreas Wieser}
	\institute{Caifa Zhou \& Andreas Wieser \at {Institute of Geodesy \& Photogrammetry}, ETH Z\"urich, Stefano-Franscini-Platz 5, 8093 Z\"urich \\ \email{caifa.zhou@geod.baug.ethz.ch, andreas.wieser@geod.baug.ethz.ch}}
	%
	%
	\maketitle

	\abstract*{We propose an approach to reduce both computational complexity and data storage requirements for the online positioning stage of a \acf{fips} by introducing segmentation of the \acf{roi}
	into sub-regions, sub-region selection using a modified Jaccard index, and feature selection based on a randomized \acf{lasso}. We implement these steps into a Bayesian framework of position estimation using the \acf{map} principle. An additional benefit of these steps is that the time for estimating the position, and the required data storage are virtually independent of the size of the \acs{roi} and of the total number of available features within the \acs{roi}. Thus the proposed steps facilitate application of \acs{fips} to large areas. 
	Results of an experimental analysis using real data collected in an office building using a Nexus 6P smart phone as user device and a total station for providing position ground truth corroborate the expected performance of the proposed approach. The positioning accuracy obtained by only processing 10 automatically identified features instead of all available ones and limiting position estimation to 10 automatically identified sub-regions instead of the entire \acs{roi} is equivalent to processing all available data. In the chosen example, 50\% of the errors are less than 1.8\ m and 90\% are less than 5\ m. However, the computation time using the automatically identified subset of data is only about 1\% of that required for processing the entire data set.}
	
	\abstract{We propose an approach to reduce both computational complexity and data storage requirements for the online positioning stage of a \acf{fips} by introducing segmentation of the \acf{roi}
	into sub-regions, sub-region selection using a modified Jaccard index, and feature selection based on randomized \acf{lasso}. We implement these steps into a Bayesian framework of position estimation using the \acf{map} principle. An additional benefit of these steps is that the time for estimating the position, and the required data storage are virtually independent of the size of the \acs{roi} and of the total number of available features within the \acs{roi}. Thus the proposed steps facilitate application of \acs{fips} to large areas. 
	Results of an experimental analysis using real data collected in an office building using a Nexus 6P smart phone as user device and a total station for providing position ground truth corroborate the expected performance of the proposed approach. The positioning accuracy obtained by only processing 10 automatically identified features instead of all available ones and limiting position estimation to 10 automatically identified sub-regions instead of the entire \acs{roi} is equivalent to processing all available data. In the chosen example, 50\% of the errors are less than 1.8\ m and 90\% are less than 5\ m. However, the computation time using the automatically identified subset of data is only about 1\% of that required for processing the entire data set.}

	\section{Introduction}
	\label{sec:1}
	Fingerprinting-based indoor positioning systems (\acsp{fips}) 
	are attractive for providing location of users or mobile assets because they can exploit signals of opportunity and infrastructure already existing for other purposes. They require no or little extra hardware, \cite{he2016wi}, and differ in that respect from many other approaches to indoor positioning like e.g., the ones using infrared beacons \cite{1432143}, ultrasonic signals \cite{hazas2006broadband}, \acf{rfid} tags \cite{bekkali2007rfid}, \acf{uwb}
	signals \cite{ingram2004ultrawideband}, or foot-mounted \acfp{imu} \cite{Gu2017}. \acs{fips} benefit from the spatial variability of a wide variety of observable features or signals like \acf{rss} from \acf{wlan} \acfp{ap}, magnetic field strengths, or ambient noise levels. \acs{fips} are therefore also called feature-based indoor positioning systems \cite{6597797}. The attainable quality of the position estimation using \acs{fips} mainly depends on the spatial gradient of the features and on their stability or predictability over time \cite{Ndrmyr2014}.   
	
	Key challenges of \acs{fips} are discussed e.g., in \cite{kushki2007kernel} and more recently in \cite{he2016wi}. The former publication focuses on four challenges of \acs{fips} utilizing vectors of \acs{rss} from \acs{wlan} \acs{ap} as fingerprints. In particular, the paper addresses i) the generation of a fingerprint database to provide a \acf{rfm} for positioning, ii) pre-processing of fingerprints for reducing computational complexity and enhancing accuracy, iii) selection of \acsp{ap} for positioning, and iv) estimation of the distance between a fingerprint measured by the user and the fingerprints represented within in the reference database. Extensions to large indoor regions and handling of variations of observable features caused by the changes of indoor environments or signal sources of the features (e.g., replacement of broken \acsp{ap}) are addressed in \cite{he2016wi}.
	
	Two widely used fingerprinting-based location methods, which we also employ herein are, $ k $-nearest neighbors ($ k $NN) \cite{Padmanabhan2000} and \acf{map} \cite{Youssef2008}. The time and storage computational complexity of both methods is proportional to the number of reference locations in the \acs{rfm} (i.e. the area of the \acs{roi}) and the number of observable features. This means that these approaches become computationally expensive in large \acsp{roi} with many \acsp{ap}.

	The goal of this paper is to propose three steps in order to facilitate accurate and flexible indoor positioning in a large \acf{roi} with potentially very high numbers of available features and feature availability varying across the \acs{roi}. For flexibility e.g., with respect to including different types of features, performing quality prediction of estimated locations, and performing quality control of measured and modeled feature values, we choose a Bayesian approach to position estimation using the \acf{map} principle. The three steps proposed herein are intended to reduce the computational complexity in terms of processing time and storage requirements, in particular during the position estimation stage which may either have to be carried out for a single user on the mobile device or for a potentially large number of users concurrently on a server. Furthermore, the steps help to make the computational effort for position estimation almost independent of the size of the \acs{roi} and of the total number of observable features within the \acs{roi}, which are important aspects for application of \acs{fips} in large areas.
	
	The first step is the segmentation of the entire \acs{roi} into non-overlapping sub-regions. The next step is the identification of approximations of the user location with a granularity corresponding to the size of the sub-regions such that the actual position estimation can be restricted to a search or optimization within a few candidate sub-regions. We apply a modified Jaccard index, \cite{park2010growing,Jani2015}, within this step, see \secPref\ref{subsubsec:modifedJin}. The final step is the identification of relevant features within each sub-region and the subsequent selection of a small number of relevant features available both in the measured fingerprint and in the \acs{rfm} for the actual position estimation. We base this feature selection on a randomized \acf{lasso} approach, \cite{tibshirani1996regression}, see \secPref\ref{subsubsec:randomizedLasso}.

	\section{Related work}\label{sec:related}
	
	\subsection{Sub-region selection}\label{subsec:subregion}
	There are mainly two types of approaches for sub-region selection\footnote{In other publications, sub-region selection is called spatial filtering \cite{kushki2007kernel}, location-clustering \cite{youssef2003wlan}, or coarse localization \cite{feng2012received}.}: approaches based on clustering and approaches based on similarity metrics. \cite{feng2012received} and \cite{chen2006power} applied affinity propagation and a $k$-means algorithm, respectively, to divide the \acs{roi} into a given number of sub-regions according to the features collected within the \acs{roi}. Both papers present clustering-based sub-region selection and require prior definition of the desired number of sub-regions and knowledge of all features observable within the entire \acs{roi}. These clustering-based approaches take the fingerprint measured by the user into account during the clustering process which may thus have to be repeated with each new user fingerprint obtained.
	
	Similarity metric-based sub-region selection instead identifies the sub-region whose fingerprints contained in the \acs{rfm} are most similar to the fingerprint observed by the user. They differ depending on the chosen similarity metric. E.g., \cite{kushki2007kernel} use the Hamming distance for this purpose, measuring only the difference in terms of observability of the features, not their actual values. Still, these approaches typically need prior information on all observable features within the entire \acs{roi} when associating a user observed fingerprint with a sub-region. This may be a severe limitation in case of a large \acs{roi} or changes of availability of the features. Modified Jaccard index-based sub-region selection as used in this paper belongs to the latter category. However, the approach proposed herein requires only the prior knowledge of the features observable within each sub-region when computing the similarity metric between the observations in the \acs{rfm} and in the observed user fingerprint. 
	
	\subsection{Selection of relevant features}\label{subsec:figerprintSelection}
	Approaches to selection of features actually used for positioning differ \acs{wrt} several perspectives. We focus on three: i) whether they take the relationship between positioning accuracy and selected features into account, ii) whether they help to reduce the computational complexity of position estimation, and iii) whether they are applicable to a variety of features or only features of a certain type. The chosen features for positioning should be the ones allowing to achieve the best positioning accuracy using the specific fingerprinting-based positioning method or achieving a useful compromise between accuracy and reduced computational burden. 
	
	Previous publications focused on feature selection for \acs{fips} using \acs{rss} from \acs{wlan} \acsp{ap} and consequently addressed the specific problem of \acs{ap} selection rather than the more general feature selection. \cite{chen2006power} and \cite{feng2012received} proposed using the subsets of \acsp{ap} whose \acs{rss} readings are the strongest assuming that the strongest signals provide the highest probability of coverage over time and the highest accuracy. \cite{kushki2007kernel} and \cite{chen2006power} applied a divergence metric (Bhattacharyya distance and information gain, respectively) to minimize the redundancy and maximize the information gained from the selected \acsp{ap}. The limitations of these approaches are: i) they are only applicable to the \acs{fips} based on \acs{rss} from \acs{wlan} \acsp{ap}, and ii) they only take the values of the features into account as selection criteria instead of the actual positioning accuracy. \cite{5210101} proposed an \acs{ap} selection strategy able to choose \acsp{ap} ensuring a certain positioning accuracy using a nonparametric information filter. However, this approach uses continuously measured fingerprints to select the subset of \acsp{ap} maximizing the discriminative ability \acs{wrt} localization. This method therefore needs several online observations for estimating one current position.
	
	In this paper, we propose an approach based on randomized \acs{lasso} to choose the most relevant features for fingerprinting-based positioning. This method differs from previous ones in three ways: i) it takes the positioning error into account,
	ii) the feature selection can be pre-computed and thus allows reducing the computational complexity of position estimation, and iii) it is a general feature selection method applicable also to fingerprints containing different types of features simultaneously.

	\section{The proposed approach}\label{sec:proposedApproach}
	In this section, we briefly summarize the fundamentals of fingerprinting-based positioning and present the main contributions of this paper, contributing to reduced computational complexity independent of the size of the \acs{roi}. In particular we present  i) candidate sub-region selection using a modified Jaccard index, ii) selection of relevant features using randomized \acs{lasso}, and iii) \acs{map}-based positioning benefiting from the previous two steps. Finally we briefly discuss the computational complexity of the proposed method. 
	\subsection{Problem formulation}\label{subsec:problemFormulation}
	Generally, an \acs{fips} is realized using two stages: offline and online stage {(Fig. 1)}. The result of the former is the \acf{rfm}, i.e., a model representing the relation between the observable features and location. At the online stage, the user's location is estimated by matching the currently measured fingerprint to the \acs{rfm} using a fingerprinting-based positioning method (e.g., \acf{map}-based positioning).
	\begin{figure}[!b]
		\centering
		\includegraphics[width=\columnwidth]{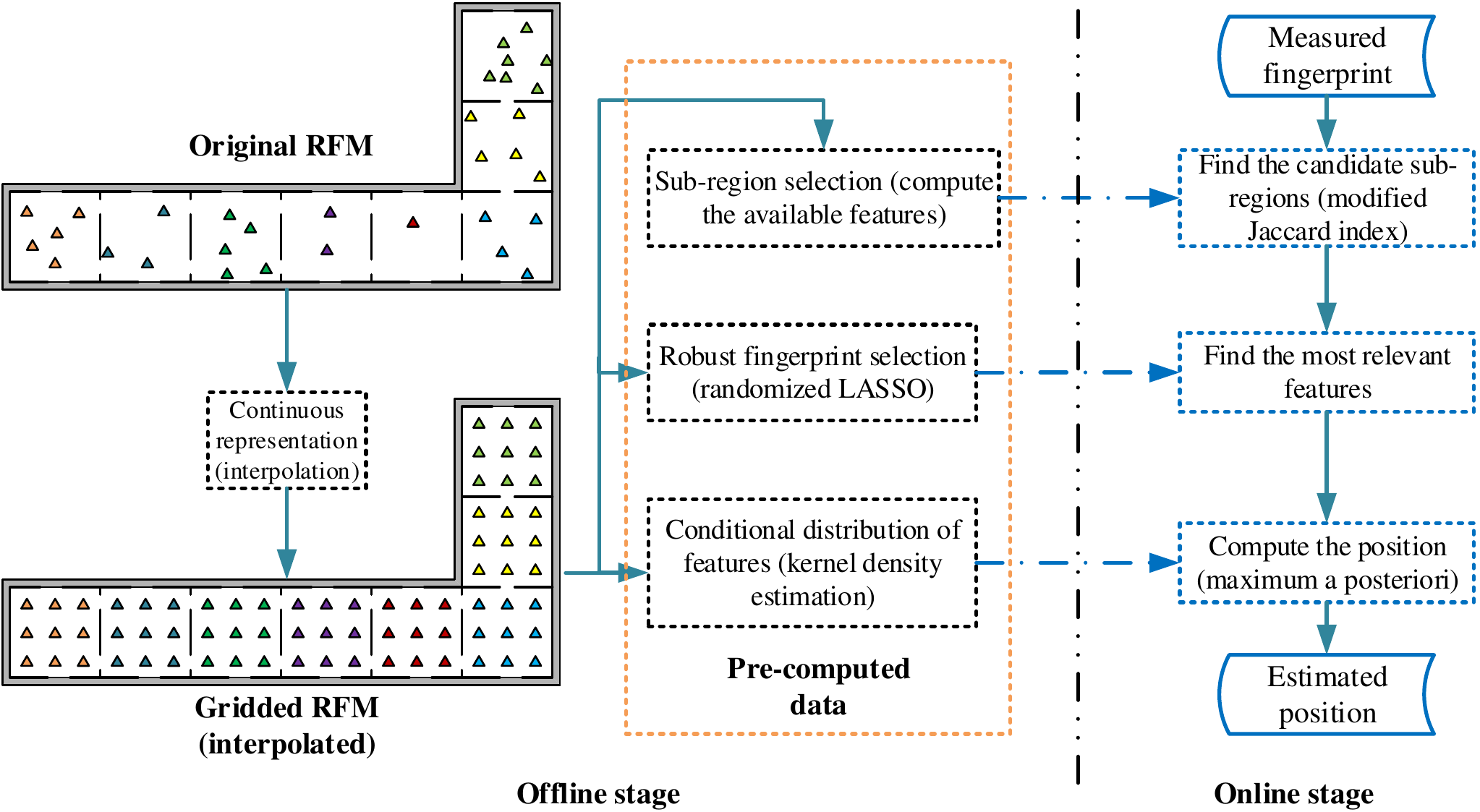}
		%
		%
		\caption{The proposed framework. In order to make the number of reference points within all sub-regions equal we interpolate the reference data in the original \acf{rfm} to provide a denser regular grid of reference points. This interpolated \acs{rfm} is used to calculate the pre-computed data for online positioning.}
		\label{fig:schematic}       
	\end{figure}

	We chose a representation herein where the \acs{rfm} is a discrete set of fingerprints associated with chosen reference positions throughout the \acf{roi}. Each fingerprint is an associative array consisting of a collection of (key, value) pairs. The key is a unique identifier of the respective feature, e.g., in case of WiFi-based fingerprinting an integer obtained by hashing the \acf{mac} address of an \acs{ap}. In case of a measured fingerprint the value is the measurement of the corresponding feature. In case of a fingerprint within the \acs{rfm} the value is the expected value of the feature at the corresponding reference position.
	
	Herein we assume that the measurements for establishing the \acs{rfm} have been made at arbitrary locations across the \acs{roi} and the \acs{rfm} is obtained by spatial interpolation using nearest-neighbor \cite{Watson1984} to obtain the fingerprints at a regular gird of reference points. Further details are given in \secPref\ref{sec:experimentalResults}. 
	
	For reducing the computational effort during the online stage, we propose to divide the \acs{roi} into $N$ non-overlapping grid cells (sub-regions) $g^i$ such that $ \mathrm{RoI}=\{g^1, g^2, \cdots, g^N\} $. In the 2D case each sub-region can simply be a rectangle or square in the coordinate space \footnote{An analysis of strategies for optimum definition of the sub-regions in terms of size and shape is left for future work.}.  Let there be $ M^{i} $ fingerprints $ \mathbf{O}^{ij}\in\mathbb{R}^{L^{ij}\times2} $ within the $i^{\mathrm{th}}$ grid cell of the \acs{rfm} where $j\in\{1,\,2,\,\dots,\,M^i\}$ and  $L^{ij}$ is the number of features observed or observable at the corresponding location. Each of these fingerprints is associated with a position $ \mathbf{l}^{ij}\in\mathbb{R}^{D} $ where $ D $ is the dimension of the coordinate space. Later on, we use the symbols $ \mathbf{o}_{\mathrm{keys}}^{ij} $ and $ \mathbf{o}_{\mathrm{values}}^{ij} $ to represent the vectors of keys and values separately such that $\mathbf{O}^{ij} = (\mathbf{o}_{\mathrm{keys}}^{ij},\,\mathbf{o}_{\mathrm{values}}^{ij})$. For arguments where the sequence of the elements is irrelevant we will later use the same symbols to also indicate the sets of keys and values with $ |\mathbf{o}_{\mathrm{keys}}^{ij}|= |\mathbf{o}_{\mathrm{values}}^{ij}|= L^{ij}$, where $ |\cdot| $ denotes the cardinality of a set.
	
	At the online stage a newly measured fingerprint $ \mathbf{O}^{\mathrm{u}}\in\mathbb{R}^{L^{\mathrm{u}}\times 2} $ becomes available at the unknown user location $ \mathbf{l}^{\mathrm{u}} $ where $ L^{\mathrm{u}} $ denotes the number of observed features at this location. This fingerprint is also represented by keys and values, i.e. $ \mathbf{O}^{\mathrm{u}} =(\mathbf{o}_{\mathrm{keys}}^{\mathrm{u}}, \mathbf{o}_{\mathrm{values}}^{\mathrm{u}})$. The positioning process consists in inferring the estimated user location $ \hat{\mathbf{l}}^{\mathrm{u}}=f(\mathbf{O}^{\mathrm{u}}) $ as a function of the fingerprint and the \acs{rfm} where $f$ is a suitable mapping from fingerprint to location, i.e. $ f: \mathbf{O}\mapsto \mathbf{l}$. We subsequently focus on the following proposed solutions to mitigate the computational load associated with offline and online stage:
	\begin{itemize}
		\item selecting the sub-region as a coarse approximation of the actual user location based on a modified Jaccard index; 
		\item identifying the most relevant features within each grid cell using the randomized \acf{lasso} algorithm such that the actual location calculation can later be carried out using only those instead of using all features;
		\item combining the above two steps with a \acf{map}-based positioning approach and implementing it in a way to keep the computational complexity of the online stage almost independent of the size of the \acs{roi} and of the total number of observable features within the \acs{roi}.  
	\end{itemize}
	
	\subsection{Sub-region selection using modified Jaccard index}\label{subsubsec:modifedJin}
	By analyzing the similarity between the keys of the measured fingerprints and the keys associated with the individual sub-regions in the \acs{rfm} we can identify candidates of sub-regions most likely containing the actual user location. The subsequent estimation of the user location can then be limited to the selected candidate regions thus reducing the computational load assuming that the index calculation is computationally less expensive than the position estimation. We use a modified Jaccard index $ c(g^i, \mathbf{O}^{\mathrm{u}})\in \left[0,1\right] $, \cite{Jani2015}, as the similarity metric. It is calculated for the observed fingerprint $ \mathbf{O}^{\mathrm{u}} $ and each sub-region by:
	\begin{equation}
		\label{eq:modifiedJC}
		c(g^i, \mathbf{O}^{\mathrm{u}}) = \frac{1}{2}\Big(\frac{|\mathbf{o}^i_{\mathrm{keys}}\cap \mathbf{o}^{\mathrm{u}}_{\mathrm{keys}}|}{|\mathbf{o}^i_{\mathrm{keys}}\cup \mathbf{o}^{\mathrm{u}}_{\mathrm{keys}}|} + \frac{|\mathbf{o}^i_{\mathrm{keys}}\cap \mathbf{o}^{\mathrm{u}}_{\mathrm{keys}}|}{|\mathbf{o}^{\mathrm{u}}_{\mathrm{keys}}|}\Big)
	\end{equation}
	Here, $ \mathbf{o}^i_{\mathrm{keys}}\in\mathbb{R}^{L^i} $ is the set of unique keys representing the observable features within the $ i^{\mathrm{th}} $ grid cell, i.e., the union of the keys ($ \mathbf{o}_{\mathrm{keys}}^{ij} $ now considered as sets) of all fingerprints within this cell:
	\begin{equation}
		\label{eq:keySets}
		\mathbf{o}^i_{\mathrm{keys}} = \underset{j}{{\cup}}\,{\mathbf{o}^{ij}_{\mathrm{keys}}},\,j=1,2,\cdots, M^i.
	\end{equation}
	The first term in \eqref{eq:modifiedJC} is the Jaccard index \cite{park2010growing}, which indicates the fraction of features common to the currently measured fingerprint and to the sub-region. The maximum value of 1 is obtained for this term (and the entire expression) if the features in the fingerprint are exactly all the features available within the sub-region. A lower value indicates that there are features which are missing either in the current fingerprint or in the \acs{rfm} of the sub-region. The second term in \eqref{eq:modifiedJC} is a modifier causing the index to favor sub-regions containing all or most features observed by the user over sub-regions lacking some of these features. The underlying assumption is that the user may not be able to observe all actually available features while the \acs{rfm} is nearly complete and it is therefore unlikely to observe features missing in the \acs{rfm}.
 
	The $ k $ sub-regions with the highest values of the modified Jaccard index are selected as candidate sub-regions for the subsequent positioning. Their cell indices are collected in the vector $ \mathbf{s}^{\mathrm{u}}_k \in\mathbb{N}^k$ for further processing. If the sub-regions are non-overlapping, as introduced above, $k$ needs to be large enough to accommodate situations where the actual user location is close to the border between certain sub-regions and small enough to reduce the computational burden of the subsequent user location estimation. We will further discuss this in \secPref\ref{sec:experimentalResults}.

	\subsection{Feature selection using randomized \acs{lasso}}\label{subsubsec:randomizedLasso}

	In a real-world environment there may be a large number of features available for positioning, e.g., hundreds of \acsp{ap} may be visible to the mobile user device in certain locations. Not all of them will be necessary to estimate the user location. In fact using only a well selected subset of the available signals instead of all may provide a more accurate estimate and will reduce the computational burden. Furthermore the number of observable features typically varies across the \acs{roi} e.g., due to Wifi \acs{ap} antenna gain patterns, structure and furniture within a building. However, it is preferable to use the same number of features throughout the candidate sub-regions for assessing the similarity between the measured fingerprint and the ones extracted from the \acs{rfm} during the on-line phase.
	
	We therefore recommend selecting a fixed number $h$ 
	of features per candidate sub-region for the final position estimation. To facilitate this selection during the on-line phase, the relevant features within each sub-region are already identified beforehand once the \acs{rfm} is available. We use an approach based on randomized \acs{lasso}, an $ L_1 $-regularized linear regression model \cite{tibshirani1996regression},
	for this step. Each feature within the sub-region is associated with an estimated coefficient by this approach. If the coefficient is sufficiently different from zero the corresponding feature is identified as relevant. During the on-line phase $h$ features (possibly different for each sub-region) are selected among the identified ones such that they are available both within the \acs{rfm} and the user fingerprint. 

	The total number of observable features within the $i^{\mathrm{th}}$ subregion is $ |\mathbf{o}^i_{\mathrm{keys}}|=L^i $. 
	To represent all fingerprints of this sub-region in the \acs{rfm} by vectors of the same dimension we replace each $\mathbf{o}^{ij}_{\mathrm{keys}}$ by $\mathbf{o}^i_{\mathrm{keys}}$, see \eqref{eq:keySets}, and the corresponding vector of values $\mathbf{o}^{ij}_{\mathrm{values}}$ by a vector $\mathbf{f}^{ij}\in\mathbb{R}^{L^i}$ which contains just the corresponding element from $\mathbf{o}^{ij}_{\mathrm{values}}$ for each feature whose key is in $\mathbf{o}^{ij}_{\mathrm{keys}}\cap\mathbf{o}^i_{\mathrm{keys}}$. For all other features it contains a value indicating that the feature is missing (e.g., a value lower than the minimum observable value of the corresponding feature).
	
	Feature selection using \acs{lasso} is based on estimating the coefficients $\mathbf{P}^i\in\mathbb{R}^{L^i\times D}$ of a linear regression of position  onto features according to:

	\begin{equation}
		\label{eq:lassoArgmin}
		\hat{\mathbf{P}^i} = \underset{\mathbf{P}^i}{\arg\min}\frac{1}{M^i}\sum_{j=1}^{M^i}\|{\mathbf{P}^i}^{\mathrm{T}}\mathbf{f}^{ij} - \mathbf{l}^{ij}\|_2^2 + \lambda\|\mathbf{P}^i\|_1
	\end{equation}
	where $ \lambda $ is a hyperparameter which needs to be set appropriately, and the $L_1$-norm term on the right hand side is used for regularization. Any zero element within $ \mathbf{P}^i $ indicates that the corresponding features does not contribute to the position estimation. Therefore, we identify the rows of $\hat{\mathbf{P}^i}$ whose absolute values exceed a given threshold (e.g., $ 10^{-4} $) and consider the corresponding features relevant. Their keys are collected in the vector $\mathbf{q}^i $.
	
	However, the results are affected by the choice of $\lambda$ and the optimum choice depends on the data. So, feature selection based on \acs{lasso} with any fixed priorly chosen value $\lambda$ is unstable \cite{fastrich2015constructing}. {In order to get an appropriate fixed value of $ \lambda $, we use cross validation.} {However, the stability of LASSO-based feature selection can be improved} by repeating the above process several times (e.g., 200 times) using a randomly sampled subset of fingerprints from the respective sub-region each time and finally taking the features most frequently contained in $\mathbf{q}^i $ as the actually most relevant ones. This approach is called randomized \acs{lasso} \cite{RSSB:RSSB740,wang2011random}. Although the computational cost of this process increases with increasing of size of the dataset (number of RPs and features, thus size of the \acs{roi}), it needs to be carried out only once at the offline stage. If need be, it can be implemented on a powerful computer and using parallel programming \footnote{For the dataset used in \secPref\ref{sec:experimentalResults}, it took about 64 mins for one randomization on a Windows 10 PC with 6 cores Intel Xeon CPU, 32G RAM.}. \tabPref\ref{tab:randomLasso} displays its realization for the present application in terms of pseudocode, where  $\bar{\mathbf{q}}^i $ represents the finally chosen vector of relevant keys of the $i^\mathrm{th}$ sub-region. 
	
	\begin{table}[!t]
		\centering
		\caption{Pseudocode of randomized \acs{lasso}}
		\label{tab:randomLasso}
		\begin{tabular}{ll}
			\hline
			& \textbf{Algorithm}: randomized \acs{lasso}\\
			\hline
			1: & \shortstack[l] {Input: $ \mathbf{Data} $ = \{$ \mathbf{f}^{ij} $, $ \mathbf{l}^{ij} $\} $ j=1,2,\cdots,M^i $; sampling ratio $ \epsilon \in (0, 1)$; \\ $\qquad\quad$number of randomizations $ T \in \mathbb{N} $; threshold $ h \in \mathbb{N} $} \\
			2: & Output: relevant features $ \bar{\mathbf{q}}^i $ of $ i^{\mathrm{th}} $ grid\\
			3: & for $ t =1,2,\cdots, T $:\\
			4: &$\qquad$$\tilde{\mathbf{Data}} $ = sampling with replacement from $\mathbf{Data}$ with ratio $ \epsilon $\\
			5: &$\qquad$$ \mathbf{q}^i_t $ = \acs{lasso}-based fingerprint selection using $\tilde{\mathbf{Data}} $\\
			6: & end for\\
			7: & computing the frequency of selection of each feature according to $ \mathbf{q}^i_t, t=1,2,\cdots,T $\\
			8: & return $ \bar{\mathbf{q}}^i $: set of features selected most frequently\\   
			\hline
		\end{tabular}
	\end{table} 
	
	

	\subsection{\acs{map}-based positioning}\label{subsubsec:map}
	Given an \acs{rfm}, i.e. a database of fingerprints and associated reference positions, the aim of positioning is to infer the most probable location $\hat{\mathbf{l}}^{\mathrm{u}}$ of the user according to the fingerprint $ \mathbf{O}^{\mathrm{u}} $ observed at the actual but unknown location ${\mathbf{l}}^{\mathrm{u}}$. We use a variety of discrete candidate locations $ \mathbf{l} $ and apply Bayes' rule to compute for each of them the degree of belief in the assumption that the current location of the user is $ \mathbf{l} $ given the available \acs{rfm} and the currently observed fingerprint. This is an  \acs{map}-based positioning method as proposed, e.g., by \cite{park2010growing,1498348}. 
	
	The posterior probability $ \mathrm{Prob}(\mathbf{l}|\mathbf{O}^{\mathrm{u}}) $ of being at location $ \mathbf{l} $ is computed by:
	\begin{equation}
		\label{eq:generalBayesRule}
		\mathrm{Prob}(\mathbf{l}|\mathbf{O}^{\mathrm{u}})=\frac{\mathrm{Prob}(\mathbf{O}^{\mathrm{u}}|\mathbf{l})\mathrm{Prob}(\mathbf{l})}{\mathrm{Prob}(\mathbf{O}^{\mathrm{u}})}
	\end{equation}
	where $ \mathrm{Prob}(\mathbf{O}^{\mathrm{u}}|\mathbf{l}) $ is the conditional probability of the fingerprint given the assumed location $ \mathbf{l} $, and $ \mathrm{Prob}(\mathbf{l}) $ and $ \mathrm{Prob}(\mathbf{O}^{\mathrm{u}}) $ are the prior probabilities of location and fingerprint respectively. Since the prior probability of the fingerprint is independent of the candidate location the \acs{map} estimate can be obtained from:
	\begin{equation}
		\label{eq:simpleMAP}
		\hat{\mathbf{l}}^{\mathrm{u}} = \underset{\mathbf{l}}{\arg\max}\left[\mathrm{Prob}(\mathbf{O}^{\mathrm{u}}|\mathbf{l})\mathrm{Prob}(\mathbf{l})\right]
	\end{equation}
	Assuming that the observable features are conditionally independent of each other \eqref{eq:simpleMAP} can be represented by the na\"ive Bayes model:
	\begin{equation}
		\label{eq:naiveMAP}
		\hat{\mathbf{l}}^{\mathrm{u}} = \underset{\mathbf{l}}{\arg\max}\left[\prod_{j=1}^{L^{\mathrm{u}}}{\mathrm{Prob}(\mathbf{O}_j^{\mathrm{u}}|\mathbf{l})}\mathrm{Prob}(\mathbf{l})\right]
	\end{equation}
	where $ \mathbf{O}_j^{\mathrm{u}} $ denotes the $ j^{\mathrm{th}} $ observed feature at the current location. In this paper, we introduce sub-region and relevant feature selection into \acs{map}-based positioning. Therefore, we only take candidate locations in the chosen sub-regions and calculate the posterior using only the previously selected most relevant features.
	
	Thus, \eqref{eq:naiveMAP} is modified to be:
	\begin{equation}
		\label{eq:modifiedNaiveMAP}
		\hat{\mathbf{l}}^{\mathrm{u}} = \underset{\mathbf{l}\in g^i}{\arg\max}\left[\prod_{j=1}^{|\bar{\mathbf{q}}^i|}{\mathrm{Prob}(\mathbf{O}_j^{\mathrm{u}}|\mathbf{l})}\mathrm{Prob}(\mathbf{l})\right],\,\forall i\in \mathbf{s}_k^{\mathrm{u}}
	\end{equation}
	where $ \mathbf{s}_k^{\mathrm{u}} $ is the vector denoting the indices of the candidate sub-regions (see \secPref\ref{subsubsec:modifedJin}) and $ \bar{\mathbf{q}}^i $ is the set of selected relevant features of the $ i^{\mathrm{th}} $ sub-region (see \secPref\ref{subsubsec:randomizedLasso}). The conditional probability $ \mathrm{Prob}(\mathbf{O}_j^{\mathrm{u}}|\mathbf{l}) $, which models the density of the $ j^{\mathrm{th}} $ feature for a given location $ \mathbf{l} $, is estimated using kernel density estimation with a Gaussian kernel from the observations stored in the \acs{rfm}, see details e.g., in \cite{scott2015multivariate,kushki2007kernel}. 
	
	Prior knowledge of the user location, e.g. derived from previous estimates of user locations and a motion model, could be used to represent the prior probability $\mathrm{Prob}(\mathbf{l})$ of the locations. However, as in \secPref\ref{subsubsec:modifedJin}, we assume also now that no such prior information is available and can hence use equal probability of $\mathbf{l}$ across all candidate sub-regions such that also $\mathrm{Prob}(\mathbf{l})$ can be dropped from \eqref{eq:modifiedNaiveMAP}.

	\subsection{Computational complexity of online positioning}\label{subsubsec:comCom}
	In this part, we analyze the computational complexity of the proposed approach and compare it to \acs{map}-based positioning without sub-region and feature selection. For the latter, the computational complexity of estimating one position is $ \mathcal{O}(\alpha N (|\mathbf{o}_{\mathrm{keys}}^{\mathrm{RoI}}| + 1)) $, where $ \alpha $ is the number of candidate locations in each of the $N$ sub-regions, and $ |\mathbf{o}_{\mathrm{keys}}^{\mathrm{RoI}}| $ denotes the total number of observable features in the entire \acs{roi}:
	\begin{equation}
		\label{eq:totalAP}
		\mathbf{o}_{\mathrm{keys}}^{\mathrm{RoI}} = \underset{i}{\cup}\mathbf{o}_{\mathrm{keys}}^{i},\, i=1,2,\cdots, N.
	\end{equation}
	
	The computational complexity of the proposed method is $ \mathcal{O}(\alpha k (\underset{i\in \mathbf{s}_k^{\mathrm{u}}}{\max}\{|\bar{\mathbf{q}}^i|\}+1)) $, where $ k $ is the number of selected sub-regions and $|\bar{\mathbf{q}}^i|$ is the number of selected features. So, clearly the computational complexity of the proposed approach is significantly less than for the \acs{map}-based approach without sub-region and feature selection. Furthermore, it is independent of the size of the \acs{roi} and of the total number of available features within the \acs{roi}. {The proposed approach is to constrain and limit the search to a set of candidate reference locations and selected features for the online positioning. Though we only give the analytical formula of the computational complexity of MAP, other fingerprinting-based location methods will also benefit from the proposed approach, because the computational complexity of fingerprinting-based positioning is proportional to the size of the search space.}
	
	The computational complexity of the proposed approach can also be kept low by an appropriate implementation strategy. Besides the \acs{rfm} further data required during the online positioning stage can be precomputed already during the offline stage (\figPref\ref{fig:schematic}). This holds in particular for:
	\begin{itemize}
		\item the set $\mathbf{o}_{\mathrm{keys}}^{i}$ of available feature keys of each sub-region required for calculating the modified Jaccard index at the online stage,
		\item the set $\bar{\mathbf{q}}^i$ of relevant features of each sub-region calculated using randomized \acs{lasso},
		\item and the conditional distribution ($ \mathrm{Prob}(\mathbf{O}_j|\mathbf{l}) $) of the selected relevant features within each sub-region obtained from kernel density estimation.  
	\end{itemize}  
	
	At the online stage these pre-computed data are cached to the user device to achieve location estimation while realizing mobile positioning. The proposed preprocessing steps also reduce the required storage space for saving the cached pre-computed data because these data only need to cover the selected relevant features instead of all the features observable within the \acs{roi}.

	\section{Experimental results and discussion}\label{sec:experimentalResults}
	In this section we analyze data obtained from real measurements collected using a Nexus 6P smart phone (for \acs{wlan} \acs{rss}) and a {Leica MS50} total station (for position ground truth) within an L-shaped \acs{roi} of about 150\ $ \mathrm{m}^2 $ in an office building for {fingerprinting-based WLAN indoor positioning, in which there are 399 observable access points. The shape of the actual floorplan corresponds to the shape given in \figPref\ref{fig:schematic}}.
	
	{We use a kinematically mapped \acs{rfm} from about 2000 reference fingerprints obtained by recording data approximately every 1.5 seconds while a user walked through the \acs{roi}. The total station tracked a prism attached to the Nexus smart phone with an accuracy of about 5 mm. This approach is a compromise between the high accuracy attainable by stop \& go measurements at carefully selected and previously marked reference positions and the low extra effort of crowd-sourced \acs{rfm} data collection as outlined e.g., in \cite{6817916}.} In order to evaluate the performance of the proposed approach independently an additional test data set was collected comprising fingerprints at approximately 500 \acfp{tp} located throughout the \acs{roi}.
	
	The coordinates of the \acsp{tp} as measured by the total station were later used as ground truth for calculating the positioning error in terms of \acf{mse} of the Euclidean distance between estimated and true coordinates. Data processing according to the proposed algorithms, as outlined in \figPref\ref{fig:schematic}, was implemented in Python using the scikit-learn package \cite{scikit-learn}.
	
	\begin{figure}[!h]
		\centering
		\subfloat[]{
			\label{subfig:jaccard}
			\includegraphics[height=0.35\columnwidth]{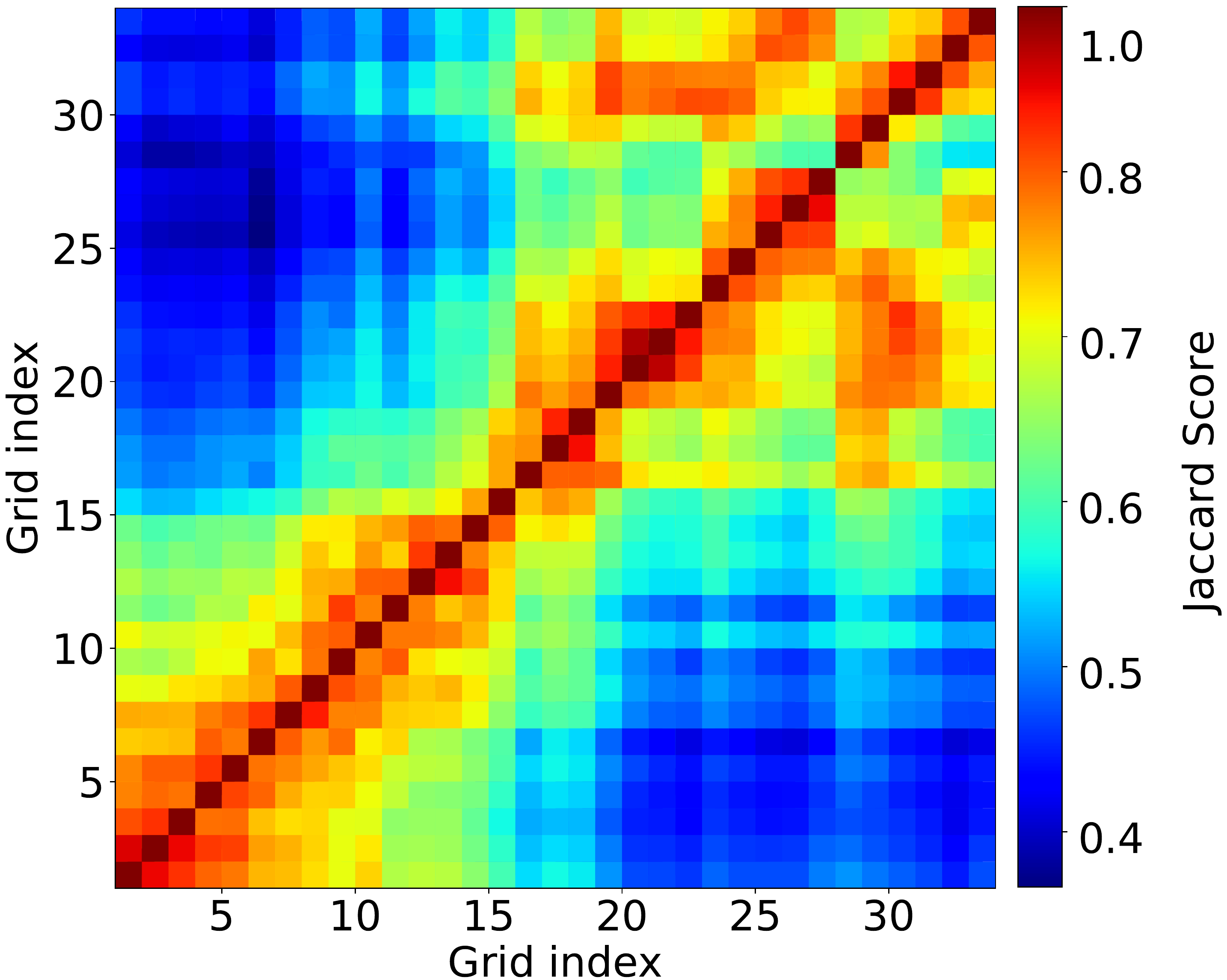}}
		\subfloat[]{
			\label{subfig:euclideandistance}
			\includegraphics[height=0.35\columnwidth]{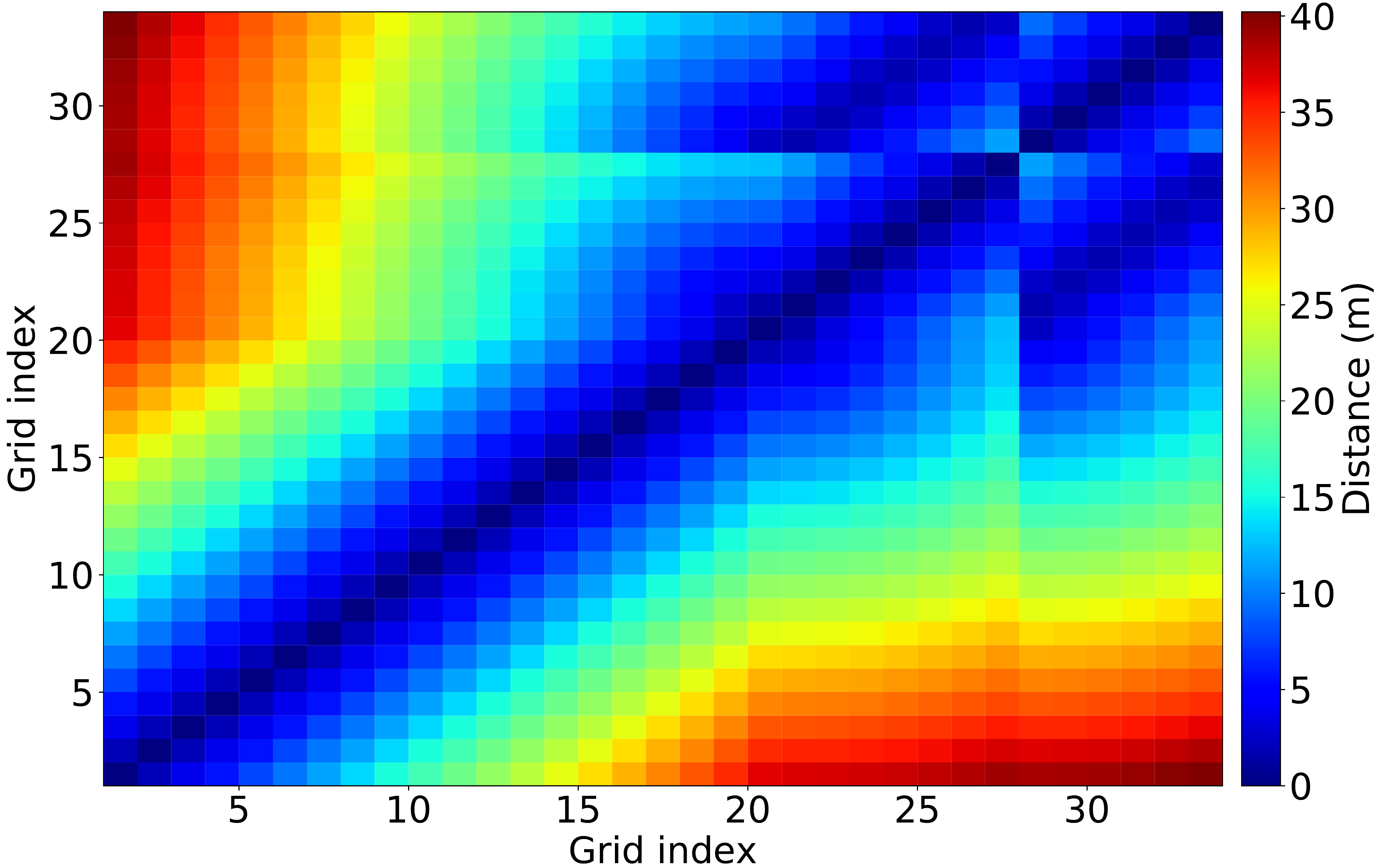}}
		\caption{Analysis of modified Jaccard index. (a): Spatial distribution of the modified Jaccard index. (b): Euclidean distance between centroids of each pair of {sub-regions}.}
		\label{fig:spatialFiltering}
	\end{figure}
	
	We divided the \acs{roi} into 34 square grid cells (sub-regions) of approximately $ 2 \times 2\, \mathrm{m}^2$ and densified the original \acs{rfm} to a regular grid of about 100 reference points per $\mathrm{m}^2$ (i.e. spacing about $ 0.2 \times 0.2\, \mathrm{m}^2$) by interpolation. The resulting gridded \acs{rfm} was used for all further processing steps.
	
	\figPref\ref{subfig:jaccard} shows the modified Jaccard index for all pairs of sub-regions indicating that the index is related to the Euclidean distance (\figPref\ref{subfig:euclideandistance}). This corresponds to the expectation that the \acs{ap}s available in nearby sub-regions are similar while different \acs{ap}s are observed in sub-regions far from each other. 
	
\begin{figure}[h]
		\centering
		\sidecaption
		\includegraphics[width=7cm]{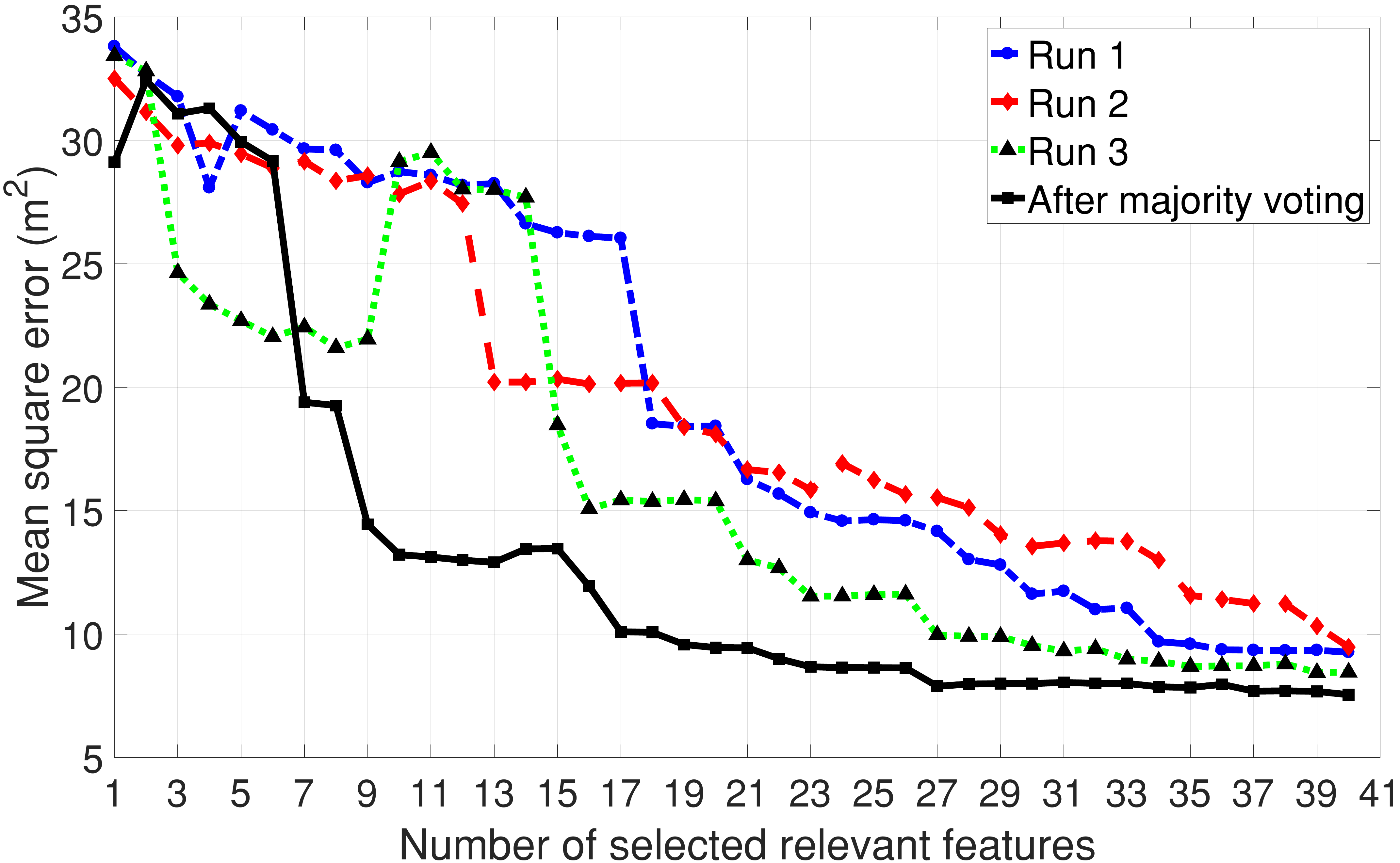}
		\caption{Empirically determined accuracy (\acs{mse}) of \acs{tp} coordinates estimated using the proposed approach with different number of selected relevant features.}
		\label{fig:msePath}
	\end{figure}

	We have then investigated the number of relevant features to be used for positioning. The keys $\bar{\mathbf{q}}^i$ of the relevant features per sub-region are the result of a randomized process and are thus random themselves. We have therefore carried out the feature selection and subsequent position estimation of the \acs{tp}s three times independently. For each of these simulation runs the \acs{mse} of the estimated coordinates of the \acsp{tp} is plotted in \figPref\ref{fig:msePath} as a function of the number $h$ of selected features actually used for positioning. {The difference of MSE of each run is caused by the randomization of LASSO-based feature selection. This figure only shows 3 out of the 200 randomized runs from which, according to \tabPref\ref{tab:randomLasso}, the final feature selection is chosen.} The figure shows that the accuracy of the estimated positions generally increases as the number of features used is increased from 1 to 20. However, the gain in accuracy is negligible if the number of features is increased above 6-10, in particular if the variability due to the randomized feature selection process is taken into account (different curves in \figPref\ref{fig:msePath}).

\begin{figure}[!h]
		\sidecaption
		\centering	
		\includegraphics[width=7.5cm]{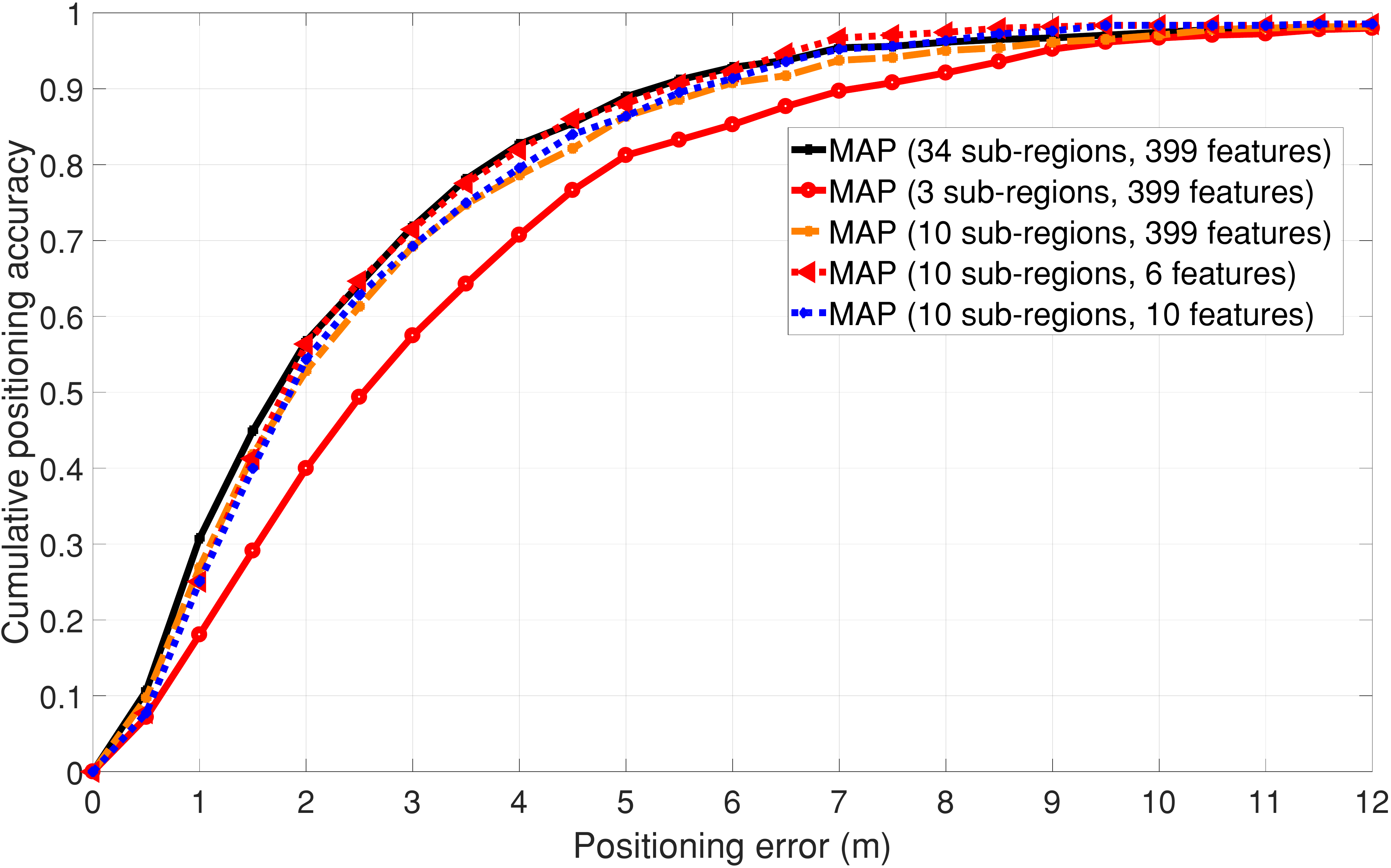}
		\caption{Empirical \acl{cpa} for different choices of parameters. The positioning error herein is the Euclidean distance between the estimated and true coordinates of the \acsp{tp}.}
		\label{fig:cpa}
	\end{figure}

	\figPref\ref{fig:cpa} illustrates the positioning accuracy and processing time for different choices of parameters within the proposed approach. The accuracy is plotted in terms of \acf{cpa} i.e., cumulative density function of the positioning errors. 
	When introducing the sub-region selection with 10 sub-regions into \acs{map}-based positioning (but using all available features), the \acs{cpa} is comparable to that of \acs{map}-based positioning without sub-region or feature selection but the average processing time \footnote{{We used Python to implement the proposed method and evaluate the processing time using the \textit{time} package (https://docs.python.org/3/library/time.html\#module-time).}} for estimating the coordinates of one \acf{tp} is 0.35 seconds (see \tabPref\ref{tab:processingTime}), which is only about 1/4 of that of \acs{map}-based positioning without sub-region or feature selection. Using only 3 sub-regions of course reduces the processing time further but leads to a considerable loss in accuracy.
	
	\begin{table}
		\caption{The processing times}
		\label{tab:processingTime}       
		%
		%
		\begin{tabular}{p{5cm}p{1.5cm}p{1.5cm}p{1.5cm}p{1.5cm}}
			\hline\noalign{\smallskip}
			\multirow{2}{*}{Methods}& \multicolumn{4}{c}{Time consumption (in \textit{s}) for positioning one \acs{tp}}  \\
			& Mean & Min & Max & Std\\
			\noalign{\smallskip}\svhline\noalign{\smallskip}
			\acs{map} (34 sub-regions, 399 features) & 1.223&1.222&1.247&0.003\\
			\acs{map} (3 sub-regions, 399 features) & 0.106&0.104&0.137&0.002\\
			\acs{map} (10 sub-regions, 399 features) & 0.353&0.347&0.388&0.003\\
			\acs{map} (10 sub-regions, 6 features) & 0.008&0.007&0.009&0.0002\\
			\acs{map} (10 sub-regions, 10 features) & 0.012&0.011&0.013&0.0002\\
			\noalign{\smallskip}\hline\noalign{\smallskip}
		\end{tabular}
	\end{table}

	By introducing both sub-region and feature selection the computational complexity and the data storage requirements can be reduced. Using 10 sub-regions and 10 features the average time of computing the coordinates of one \acs{tp} is only 1\% of that of \acs{map}-based positioning with neither sub-region nor feature selection while the attained accuracy is virtually equal to the one obtained using all data. In agreement with the results depicted in \figPref\ref{fig:msePath} there is no significant loss in accuracy when using 6 features instead of 10, while the processing time decreases roughly by a factor of 2. 
	
	\section{Conclusion}\label{sec:futureWork}
	We proposed herein an approach to fingerprinting-based indoor positioning using the \acf{map} principle for coordinate estimation. The main contributions are proposals to reduce data storage requirements and computational complexity in terms of processing times by segmentation of the entire \acs{roi} into sub-regions, identification of a few candidate sub-regions during the online positioning stage, and use of a selected subset of features instead of all available features for position estimation. Sub-region selection is based on a modified Jaccard index measuring the similarity between the features obtained by the user and those available within the \acf{rfm}. Feature selection is based on the randomized \acf{lasso} yielding a pre-computed set of relevant features for each sub-region. The reduction of computational complexity is obtained both from the reduction of the number of candidate locations needed to analyze during online positioning and from the reduction of the number of features to be compared. 
	
	The experimental results corroborated the claim of reduced complexity while indicating that the positioning accuracy is hardly reduced by processing only 10 candidate subregions instead of the entire \acs{roi} and by selecting only 6-10 features instead of using all available ones. Given a fixed number of candidate sub-regions and a fixed, low number of features the computational burden of the entire algorithm is almost independent of the size of the entire \acs{roi} and of the number of available features across the \acs{roi}.
	
	Further work will concentrate on increasing the stability of the feature selection via adaptive forward-backward greedy feature selection \cite{zhang2011adaptive}, on ranking features with respect to quality and impact, on taking into account user motion during sub-region selection and on handling temporal changes of the \acs{rfm}. Furthermore, we are currently applying the proposed approach to larger and more complex datasets \cite{montoliu2017} and migrating to a mobile phone.
	
	\begin{acknowledgement}
	{The China Scholarship Council  (CSC) financially supports the first author's doctoral research. Questions and proposals by three anonymous reviewers are acknowledged for contributing to improved quality of the paper.}
	\end{acknowledgement}
	%
	%
	
	%
	\bibliographystyle{apalike}
	\bibliography{lbs_2018}
	
\end{document}